\documentclass[letterpaper]{article} 
\usepackage{aaai23}  
\usepackage{times}  
\usepackage{helvet}  
\usepackage{courier}  
\usepackage[hyphens]{url}  
\usepackage{graphicx} 
\usepackage{natbib}  
\usepackage{caption} 
\usepackage{algorithm}
\usepackage{algorithmic}
\usepackage{multirow}
\usepackage{amsmath} 
\usepackage{amssymb}  
\usepackage[table]{xcolor}
\usepackage{svg}
\usepackage{tikz}
\usepackage{placeins}

\usepackage{soul}

\newcommand{\add}[1] {\textcolor{black}{#1}}

\usepackage{diagbox}
\usepackage{newfloat}
\usepackage{listings}
\DeclareCaptionStyle{ruled}{labelfont=normalfont,labelsep=colon,strut=off} 
\floatstyle{ruled}
\newfloat{listing}{tb}{lst}{}
\floatname{listing}{Listing}
\title{Effective Baselines for Multiple Object Rearrangement Planning in\\ Partially Observable Mapped Environments}
\author {
    Engin Tekin\textsuperscript{\rm 1},
    Elaheh Barati\textsuperscript{\rm 2}\thanks{Work done at Meta}, 
    Nitin Kamra\textsuperscript{\rm 1}, 
    and Ruta Desai\textsuperscript{\rm 1}\\
}
\affiliations {
    \textsuperscript{\rm 1} Meta Reality Labs Research, \textsuperscript{\rm 2} TikTok\\
    \{tekinengin21, nitinkamra, rutadesai\}@meta.com, elaheh.barati@gmail.com
    
}

\begin{document}

\maketitle

\begin{abstract}
Many real-world tasks, from house-cleaning to cooking, can be formulated as multi-object rearrangement problems -- where an agent needs to get specific objects into appropriate goal states. For such problems, we focus on the setting that assumes a pre-specified goal state, availability of perfect manipulation and object recognition  capabilities, and a static map of the environment but unknown initial location of objects to be rearranged. Our goal is to enable home-assistive intelligent agents to efficiently plan for rearrangement under such partial observability. This requires efficient trade-offs between exploration of the environment and planning for rearrangement, which is challenging because of long-horizon nature of the problem. To make progress on this problem, we first analyze the effects of various factors such as number of objects and receptacles, agent carrying capacity, environment layouts etc. on exploration and planning for rearrangement using classical methods. We then investigate both monolithic and modular deep reinforcement learning (DRL) methods for planning in our setting. We find that monolithic DRL methods do not succeed at long-horizon planning needed for multi-object rearrangement. Instead, modular greedy approaches surprisingly perform reasonably well and emerge as competitive baselines for planning with partial observability in multi-object rearrangement problems. We also show that our greedy modular agents are empirically optimal when the objects that need to be rearranged are uniformly distributed in the environment -- thereby contributing baselines with strong performance for future work on multi-object rearrangement planning in partially observable settings.
\end{abstract}

\section{Introduction}
Rearrangement problems, where the goal is to get a physical environment in a specific state has been proposed as the next frontier for embodied AI research~\cite{RearrangementChallange}. Many tasks in everyday life from house cleaning~\cite{kondo, szot2021habitat} to preparing groceries~\cite{szot2021habitat} can be formulated as rearrangement problems. Therefore, developing embodied agents to solve these problems would allow to us to make progress towards the next generation of home assistant agents.

In an embodied rearrangement task, an agent must rearrange an unknown environment using \add{a combination of} sensor observations \add{and prior knowledge} to reach a goal state, which is specified either geometrically, or through image, language, or predicates~\cite{RearrangementChallange}. Solving such generic rearrangement tasks requires an agent to solve a plethora of sub-problems, such as, reasoning about the goal state through semantic and commonsense understanding; building a map of the environment in order to navigate, search, and explore; effectively planning to figure out which objects to pick-drop and in what order; and finally manipulating objects. These problems span the spectrum of perception, planning, navigation, and manipulation making rearrangement an extremely challenging problem~\cite{RearrangementChallange}.

Because of the complexity of rearrangement problems, previous research has focused on different slices of the problem. Some researchers focus on understanding the goal state by leveraging human preferences and commonsense reasoning~\cite{housekeep, kapelyukh2022my}; or through reasoning about changes in the environment configuration~\cite{VisualRe, sceneChangeChallenge}. Another body of work focuses on perception, planning, and manipulation for rearrangement, albeit for small number (upto 5) of objects~\cite{szot2021habitat, tabula-rasa}. We focus on a specific slice of the rearrangement problem -- \emph{planning under partial observability}. In particular, to decouple our investigation from the manipulation, navigation and perception challenges, we assume perfect object recognition and interaction capabilities in the agent. We further assume the availability of a static map to focus on high-level task planning for rearrangement instead of integrated task and motion planning (TAMP)~\cite{garrett2021integrated}. Lastly, instead of dealing with uncertain information ranging from incomplete maps, mis-classified objects etc., which makes it challenging to study the planning problem, we directly consider uncertainty over the object locations -- decoupled from perception -- which allows us to study the implications of uncertainty on the rearrangement planning problem in a more systematic and controlled manner. Specifically, the problem requires efficient exploration of the environment in combination with well-balanced planning for rearrangement and this forms the core of our work.


Overall, given a pre-specified goal state, perfect manipulation and object recognition capabilities, and a static map of the environment; our goal is to enable efficient planning for rearrangement. Such a laser focus consequently enables us to analyze and present the effects of various factors such as number of objects, agent carrying capacity, environment layout complexity etc. on the complexity of planning. We find that higher agent capacity and larger environment layouts make the rearrangement planning problem challenging. On the other hand, counter-intuitively, higher number of objects reduces the problem complexity, since in this case \add{rearrangement for seen objects implicitly shoulders the burden of exploration for unseen objects}, thereby reducing the need for \add{additional explicit} exploration, and ultimately reducing the problem complexity.


\add{Classical planners often fail to optimally solve task planning problems in real-time~\cite{taskogprahy} and need a world model to be known.} Motivated by \add{these} limitations of the classical approaches, we next present end-to-end monolithic deep reinforcement learning (DRL) based approaches. 
We find that akin to prior works, monolithic DRL methods do not succeed at long-horizon planning problems. We then propose modular methods as competitive baselines for planning under uncertainty in rearrangement problems. Specifically, our methods investigate ways to achieve better trade-off between exploration and planning. We empirically demonstrate that approaches which plan greedily and only explore conservatively achieve this trade-off optimally when objects that need to be rearranged have uniform distribution in the environments. We hope that our analysis and baselines will provide a good starting point and benchmark for future work on rearrangement planning.


\section{Related Work}
\subsection{Object rearrangement}
Object rearrangement has been studied in robotics~\cite{ben1998practical}, for a variety of problems ranging from table-top rearrangement~\cite{ben1998practical, objcen-robcen, large-scale} to house-scale rearrangement~\cite{taskogprahy}. 
Majority of planning research for rearrangement in robotics further focuses on integrated task and motion planning~\cite{king2017unobservable, objcen-robcen, ben1998practical}. Instead, our focus is on high-level task planning for house-scale rearrangement. Recent work from Newman et al.~(2020)~\nocite{kondo} and Agia et al.~(2022)~\nocite{taskogprahy}have also focused on task planning for house-scale object rearrangement problems, however they assume full observability while we are interested in task planning for rearrangement under partial observability.

Embodied AI community, which is interested in virtual agents in addition to physical agents, has also started pushing on house-scale rearrangement problems~\cite{RearrangementChallange, szot2021habitat, VisualRe, housekeep, Kim2021LearningTE}. Because of the use of perceptual sensors, they also have to deal with partial observability while doing rearrangement planning. These works push on solving rearrangement planning in combination with perception, manipulation, and navigation for rearrangement. Consequently, they only show rearrangement for scenarios with small number of objects (upto 5) and single agent capacity. Instead, we are interested in understanding various factors that affect rearrangement planning problem under partial observability. In particular, our focus is on the trade-off between two main components of rearrangement planning: object discovery via exploration and planning for rearrangement. We next review related work for exploration and planning.

\subsection{Exploration in 3D environments}
Exploring unknown environments is a well-studied problem in robotics and state-of-the-art results are generally achieved with frontier-based exploration (FBE)~\cite{yamauchi1998frontier, Exploration1}. The overarching idea in frontier-based approaches is to define a set of frontier locations and decide the next frontier location to explore~\cite{yamauchi1998frontier}. Recent studies have employed utility functions to choose between frontier locations in combination with DRL techniques for exploration~\cite{ Exploration1, DRL-FBE}. Ramakrishnan et al.~(2020)~\nocite{EforEVE} further benchmarked various DRL approaches with several reward functions as well as FBE method for embodied exploration. As an alternative approach to enable exploration, Fukazawa et al.~(2005)~\nocite{JapanRearrangement} uses the reaction-diffusion equation on graphs and optimize a custom potential function in order to create observation points; and then visit them in an optimal order. However, most of these studies focus solely on exploration of a 3D environment with the goal of maximizing area coverage or object/landmark detection. Instead, we are interested in exploration for object discovery to enable downstream rearrangement. We take inspiration from these studies and investigate both learning-based and frontier-based exploration approaches to enable rearrangement.

\subsection{Planning for rearrangement}
End-to-end DRL approaches as well as classical approaches based on Traveling Salesman Problem (TSP) formulation have both been used for rearrangement planning~\cite{kondo, housekeep, tabula-rasa, policyskectes, szot2021habitat}. Since multiple object rearrangement is a long-horizon problem, monolithic DRL approaches for the same have only been shown to be successful on small grid maps~\cite{tabula-rasa, policyskectes}. Researchers have therefore used modular approaches~\cite{szot2021habitat, housekeep} or classical approaches based on TSP~\cite{kondo} for house-scale rearrangement planning. Based on these findings, we also model planning for rearrangement as Capacitated Vehicle Routing Problem (CVRP), a variant of TSP, and use OR-Tools~\cite{ortools} to solve the same.

\section{Problem Characterization}
\label{sec:problem}
In this section, we concretely define the Multiple Object Rearrangement Planning (MORP) problem in mapped environments with partial observability over object locations. We also detail our procedure for generating the required datasets and evaluation metrics.

\subsection{Task Definition}

\textbf{Notation:} We will use a Partially Observable Markov Decision Processes (POMDP) framework to formally define MORP. Let $s \in \mathcal{S}$ denote a state in state space $\mathcal{S}$, $o \in \mathcal{O}$ denote an observation in observation space $\mathcal{O}$, $a \in \mathcal{A}$ denote action in action space $\mathcal{A}$, $g \in \mathcal{G}$ denote a goal specification in goal space $\mathcal{G}$, and finally $\pi(a_t|o_{t-1}...o_{0},g) = Pr(a_t|o_{t-1}...o_{0},g)$ agent's goal-conditioned policy. Note that we will omit time $t$ subscript unless indicated otherwise.

\begin{figure}[h]
    \centering
    \includegraphics[width=0.9\linewidth]{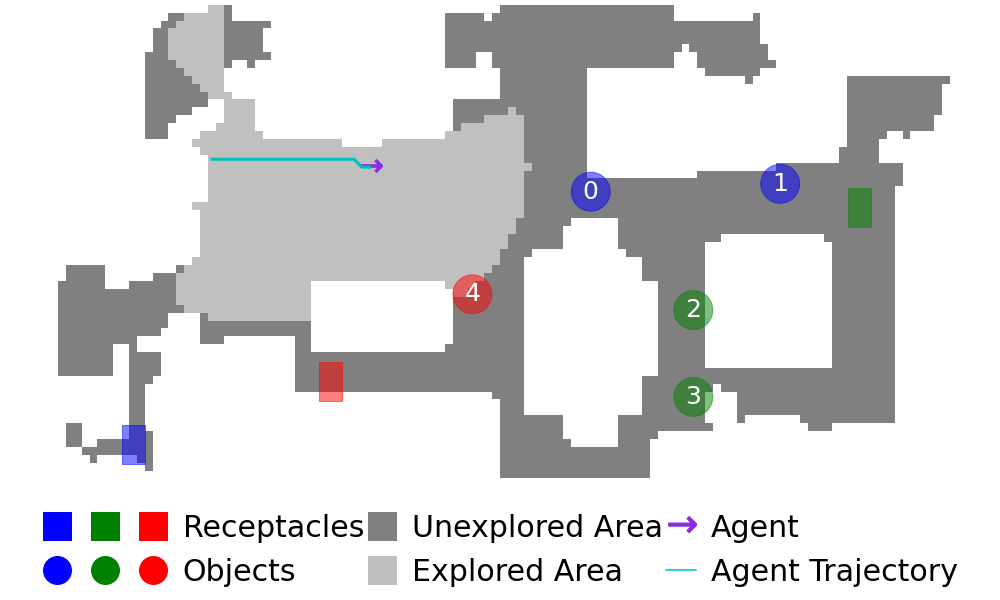}
    \caption{Multiple Object Rearrangement Planning (MORP) problem. Top down map of the house with the objects to be rearranged shown as circles and receptacles as squares. Same color signifies object-receptacle pairing and multiple objects of the same type can be placed into a receptacle (e.g., books into bookshelf). With perfect object recognition capabilities, unique instance IDs track the objects.}
    \label{fig:ROBOKONDO2D}
\end{figure}

\noindent\textbf{Goal Specification:} We consider type-level geometric goal for rearrangement as described in~\cite{RearrangementChallange} i.e., we describe the goal state for rearrangement geometrically with target location for all objects. Note that in this paper we will only use integers for location variables. Let $\mathbb{Z}$ denote the set of integers and $\mathbb{N}$ denote the set of natural numbers. We define ${p}^0_i \in \mathcal{P} : \mathbb{Z}^{n_o \times 2}$ to be initial 2D object locations and $r_i \in \mathcal{R} : \mathbb{Z}^{n_r \times 2}$ to be 2D receptacle locations where $n_o \in \mathbb{N}$ is number of the objects and $n_r \in \mathbb{N}$ number of object types/receptacles. Each object type has a unique receptacle but multiple objects can be rearranged to the same receptacle. Let $f$ be the correspondence function to map object index to an index corresponding to its receptacle type, so that we can elicit the object-receptacle pairing. A goal is then specified as $g = \bigwedge^{n_o}_{i=1}({p}_i == r_{f(i)})$.

\noindent\textbf{Rearrangement Task:} Thus, we can formally define a rearrangement task as, given a goal specification $g$, the agent must transform the initial state $s_0$ to a goal state $s^*$, by solely acting based on observations $o \in \mathcal{O}$, where $s^* \in \mathcal{S}$ is a state that satisfies goal specifications. Hence, an agent spawned randomly in a home environment must find objects scattered randomly around the house and rearrange them to their desired receptacles (see Figure~\ref{fig:ROBOKONDO2D}).

\begin{figure*}[!htbp]
    \centering
    \includegraphics[width=\linewidth]{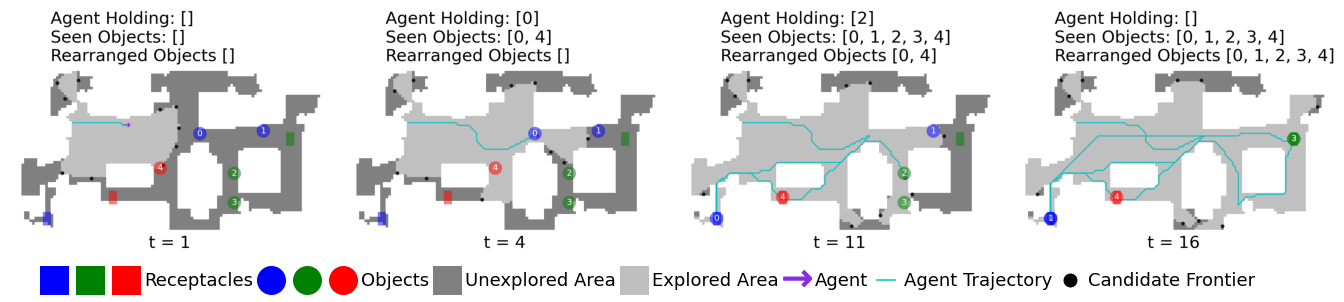}
    \caption{A sample episode for MORP -- Agent finds misplaced objects, carries them to known receptacles locations and places them. Object ids for seen and rearranged objects as well as object held by the agent are shown.}
    \label{fig:episode}
    \vspace{-10pt}
\end{figure*}

\subsection{Dataset}
\label{subsec:dataset}
\textbf{Scenes:} We choose 215 scenes from the Gibson dataset~\cite{gibsonenv}, with low clutter~\cite{hm3d} and no or limited furniture. Gibson scenes are loaded with AI Habitat simulator~\cite{ szot2021habitat} and converted to top-down occupancy maps. The top-down occupancy maps are discrete 2D grids $\mathcal{M} \in \mathbb{Z}^{200\times150}$ with elements $m_{ij} \in \{-1,0,1\} \rightarrow \{\text{Innavigable, Unexplored Navigable, Explored Navigable}\}$ with a discretization of 0.1 meter per cell. The map size of $200 \times 150$ was empirically chosen. Scene metrics are presented in Table \ref{tab:scenestats}.

\begin{table}[h]
    \centering
    \rowcolors{1}{gray!3}{gray!7}
    \begin{tabular}{c|c c c c}
        & nav-area $(m^2)$ & nav-complexity & \#scenes\\
        \hline
        large & 66.09  & 19.36 & 88 \\
        \hline
        medium & 30.53  & 12.07 & 120 \\
        \hline
        small & 1.38  & 1.39 & 7
    \end{tabular}
    \caption{Scene statistics: Navigable area\footnotemark (nav-area) and navigation complexity (nav-complexity), which is defined as the maximum ratio of geodesic and euclidean distance between any two navigable points in a scene, are shown for scenes in our large/medium/small dataset splits.}
    \label{tab:scenestats}
\end{table}
\footnotetext{We calculate navigable area using mesh calculated for agent with 1.5m height and 0.1m radius. Gibson dataset used to create top-down maps consists 1447 floors with 56 $m^2$ average navigable area. Comparing to Gibson floors our dataset consists less navigable area, since slicing a 3D scene into top-down map from a certain height does not necessarily capture whole floor area.}

\noindent\textbf{Objects and Receptacles:} We randomly place objects and receptacles on the occupancy map. Object states contain only type and location information, therefore only distinction between objects is type. Agent interaction can only modify object locations $p_i$ and there is no other parameter in the environment that changes during the episode. Objects and receptacles are not considered to occupy space on $\mathcal{M}$ to simplify collision computation. Multiple objects can be located in the same grid cell $m_{ij}$.

\noindent\textbf{Episodes:} Let $x^t:\mathbb{Z}^2$ and $\phi^t \in {\{0,..,7\}} \rightarrow{\{N, NE, E, SE, S, SW, W, NW\}}$ be location and orientation of agent at time step $t$ where \{N:North, E:East, S:South, W:West\}. At start of an episode, the agent is spawned at a random location $x^0$ and a random heading $\phi^0$ (encoded as a 8-dimensional one-hot vector) as shown in Figure \ref{fig:ROBOKONDO2D}. The episode is terminated: (a) successfully, if all objects are in their receptacles before the time-step $max_{t}$, or (b) unsuccessfully, when the maximum episode time-step ${max}_{t}$ is reached. Some objects may already be in their receptacles when the episode starts, but the agent must still verify this by at least seeing them in their receptacles for successful completion of goal $g$. Figure \ref{fig:episode} shows a successful episode completion. We split the dataset into small/medium/large categories based on the area enclosed by the map. Each category contains train/validation/test splits. Episode statistics are presented in Table \ref{tab:episodestats} in the Appendix.

\subsection{Evaluation}
\label{subsec:evaluation}
For MORP, we are interested in measuring the agent's success rate in completing the task and the agent's efficiency, i.e., distance traveled/time taken for the task. In addition, to evaluate exploration methods, we use object discovery and map coverage metrics. We describe these metrics below: \\
\textbf{Rearrangement}:
\begin{itemize}

\item  \textbf{Episode Success ($ES$)}: If all objects are seen and rearranged, $ES$ is 1 otherwise 0~\cite{housekeep}.

\item  \textbf{Rearranged Object Ratio ($ROR$)}: Ratio of the number of rearranged objects to the total number of objects that need to be rearranged.

\item  \textbf{Episodic Success Weighted by Path Length ($ESPL$)}: In order to measure agent's efficiency, we compare path length and $ES$ with oracle agent's path length. $ESPL$ is defined as 
\begin{equation}
    ESPL = ES  \frac{z}{max(z, l)}
\end{equation}
where $z \in \mathbb{R}$ is the oracle agent's path length, and $l \in \mathbb{R}$ is the path length~\cite{metrics}. An oracle agent is essentially an agent that has full observability of the environment and uses CVRP planner to optimally perform rearrangement, similar to~\cite{kondo}. We will describe this agent in more detail in Sec.~\ref{sec:benchmark}.
\end{itemize}
\textbf{Exploration}:
\begin{itemize}

\item  \textbf{Seen Object Ratio ($SOR$)}: Ratio of the number of seen objects to the total number of objects.

\item  \textbf{Map Coverage ($MC$)}: Ratio of the explored area to the total navigable area.
\end{itemize}

\subsection{Agent Definition}
\label{subsec:agent_def}
Next, we define an agent, which is capable of sensing the environment and objects, navigating around, and rearranging objects.

\vspace{3pt}
\noindent\textbf{Sensor Suite:}
\begin{itemize}
    \item Top-down Occupancy Map $\mathcal{M}$: We assume that a static top-down occupancy map of the environment $\mathcal{M}$ is available to the agent. Such a map could be obtained by a one-time scan of a house and thus may be a reasonable assumption for future home-assistant agents of any form.
    \item Receptacle Locations $\mathcal{R}$: Receptacle locations on the map $\mathcal{M}$ are available to agent.
    \item Agent Location $x$ and Orientation $\phi$: Agent can detect its location, $x=$(x,y) coordinates on $\mathcal{M}$ and orientation $\phi$ in 8 discrete directions.
    \item Field of View $FOV$: Agent can explore the map area and detect objects within its field-of-view determined by the conical sector ($\theta, r_s$) where $\theta$ is angle around the orientation direction $\phi$ and $r_s$ is the cut-off distance from $x$.
    \item Gripper $\mathcal{H} \in \mathbb{R}^{c \times n_o}$: This proprioceptive sensor contains flags indicating whether object $j$ is being held by the agent's gripper slot $i$, where $c \in \mathbb{N}$ is number of the slots equivalent to agent's carrying capacity. All objects can be held by any slot and a slot can only hold one object at a time.
\end{itemize}
\textbf{Action Space:}
The agent has three discrete actions: \emph{forward}, \emph{left} turn, and \emph{right} turn for navigation and a single \emph{grab/drop} action for object manipulation. The forward action moves the agent to one of 8 immediate grid cells in the map depending on its orientation $\phi$. The left/right turn action changes $\phi$ by $45^{\circ}$ in counter-clockwise/clockwise direction respectively. The grab/drop action is similar to ``discrete object grasping"~\cite{RearrangementChallange}. 
Such action space $\mathcal{A}$ enables abstraction from downstream embodiment including parameters for continuous control of motors and allows us to focus on high-level task planning and exploration. 

\section{Complexity Analysis and Benchmarking}
\label{sec:benchmark}
In this section, we present an analysis of factors that affect the complexity of rearrangement planning. Inspired by related work, we investigate various factors for our analysis, such as agent carrying capacity $c$~\cite{RearrangementChallange}, total number of objects $n_o$~\cite{taskogprahy}, number of receptacles $n_r$, and navigable area of the scene. To that end, we introduce heuristic and oracle agents and benchmark their performance using metrics introduced in Sec.~\ref{sec:problem} on the MORP as a way to analyze the complexity of MORP.

\subsection{Oracle Agent}
In order to find an upper bound on performance for  MORP, we consider the \emph{oracle} agent with access to privileged information. Specifically, the oracle agent has full state information and knows all object locations. Such full observability enables the agent to use a Capacitated Vehicle Routing Problem (CVRP) based approach to calculate shortest path length for rearranging objects, similar to~\cite{kondo}. We describe the CVRP formulation in more detail in Appendix~\ref{CVRPFormulation}. We solve the formulated planning problem using OR-Tools \cite{ortools}. \vspace{1pt}

The oracle agent also leverages the full state information to ignore objects that are already arranged at the start of an episode. Such CVRP-based optimal path computation for only misplaced objects truly makes the oracle agent's performance an upper bound for multiple object rearrangement planning \emph{without} any uncertainty. We next extend this oracle agent with exploration capability in a heuristic manner to deal with the uncertainty of object location in our setup. 

\subsection{Heuristic Agents}
\label{heuristics}
To benchmark the performance for multiple object rearrangement planning in presence of uncertainty (MORP), we consider \emph{heuristic} agents. We leverage the intuition that MORP requires solving two sub-problems -- (a)~efficient exploration and search of the indoor environment to deal with location uncertainty of misplaced objects,~(b)~optimal path planning balanced with such object search to rearrange the misplaced objects. Based on this intuition, our heuristic agents use a modular approach that greedily combines classical optimal approaches for both of these sub-problems.\vspace{1pt}

In particular, we use CVRP-based approach similar to the oracle agent for rearrangement planning of discovered objects and frontier-based exploration (FBE) approaches and its variants for efficient exploration and object discovery, based on the evidence of FBE's performance in classical robotics applications~\cite{DRL-FBE}. The agent greedily chooses \emph{planning} if there are any seen objects that need to be rearranged, and \emph{exploration} otherwise. Our approach of such greedy combination of these exploration and planning methods is also similar to recent work on house-scale rearrangement in EAI~\cite{housekeep, JapanRearrangement}. 

One can also think of \emph{planning} and \emph{exploration} as high-level actions. Given one of these high-level actions, the agent executes a sequence of low-level navigation actions described in Sec.~\ref{sec:problem} until certain conditions are satisfied:~(1)~\emph{planning} reaches the planned location, or~(2)~\emph{exploration} reaches the target location, or~(3)~maximum high-level action distance ${max}_{dist}$ is reached. These navigation actions are computed to follow a shortest path~\footnote{The shortest path is computed using the A$^*$ algorithm~\cite{A*}.} between the agent's current location and the target \emph{planned} or \emph{exploration} location. The agent executes \emph{grab/drop} action when it reaches a planned location and replans when new object(s) are discovered. Likewise, the agent chooses to explore if no objects have been discovered or if all the discovered objects are in their goal state and Rearranged Object Ratio (ROR) as described in Sec.~\ref{sec:problem} is less than 1. Such a modular approach of executing a low-level policy toward a goal given by a high-level policy is also inspired by recent work in embodied AI~\cite{ActiveSLAM}. \vspace{1pt}

We now describe the variants of the FBE method and a random exploration approach, which we investigated for efficient exploration in our heuristic agents. Each time the agent engages the high-level action of \emph{exploration}, a target location to be explored is obtained based on one of these approaches, described below.

\subsubsection{Weighted Frontier Based Exploration}FBE computes unexplored frontiers, which are the borderline grid-cells between explored and unexplored navigable area on a map. It then chooses the closest frontier location to the agent's location as the next location to visit for exploration~\cite{FBE}. Recent variants of FBE referred to as weighted FBE (WFBE) use a utility function to determine next the frontier location to visit~\cite{DRL-FBE}. The utility function balances the potential information gain for exploration achieved by visiting a frontier location with the distance that needs to be traveled to reach the frontier location from the agent's current location. The information gain for a given frontier is defined as the sum of newly seen grid-cells along the shortest path between the agent's current location and the frontier\footnote{We find no difference in WFBE performance when the information gain is computed only at the frontier location instead.}. Inspired by the performance of WFBE method, we investigate two variants  of WFBE, which leverage two different utility functions: 

\begin{itemize}
    \item \textbf{WFBE$_\text{r}$}: uses the ratio between frontier information gain and frontier distance as the utility function. It chooses a frontier that maximizes this function.
    \item \textbf{WFBE$_\text{w}$}: uses weighted sum of normalized distance and normalized gain as the utility function, where the weight $w$ is weight on distance and $(1-w)$ is the weight on gain. It then chooses the frontier that minimizes this function. We experiment with different values of the $w$:~{$0,0.5,1$} respectively. 
\end{itemize}

Please refer to the Appendix~\ref{WFBEAppendix} for more details on the utility functions.


\subsubsection{Random Exploration (RND)}This approach picks a random navigable location from the unexplored area as the target location for exploration.


\begin{table}[h]
    \fontsize{9pt}{10pt} \selectfont
    \centering
    \rowcolors{1}{gray!3}{gray!7}
    \begin{tabular}{c|c|c|cccc||c}
        \multicolumn{1}{c}{ } &\multicolumn{1}{c}{}&\multicolumn{1}{c|}{}&
        \multicolumn{5}{c}{$ESPL$}\\
        \hline
         \textit{Explorer} & $c$ & \diagbox[innerwidth=0.5cm]{$n_r$}{$n_o$} & \multicolumn{1}{c|}{$1$} & \multicolumn{1}{c|}{$3$} & \multicolumn{1}{c|}{$5$} & \multicolumn{1}{c||}{$10$} & \multicolumn{1}{c}{\textit{avg}}\\
        \hline
        & 1 & 3 & .62 & .69 & .74 & .82 & .72\\
        \multirow{-2}{*}{RND} & 3 & 3 & .62 & .63 & .60 & .62 & .62\\
        \hline
        & 1 & 3 & .56 & .67 & .75 & .83 & .71\\
        \multirow{-2}{*}{WFBE$_{1}$} & 3 & 3 & .56 & .62 & .61 & .64 & .61\\
        \hline
        & 1 & 3 & \textbf{.68} & .75 & \textbf{.81} & \textbf{.87} & \textbf{.78}\\
        \multirow{-2}{*}{WFBE$_{0.5}$} & 3 & 3 & \textbf{.68} & .69 & \textbf{.65} & .67 & .67\\
        \hline
        & 1 & 3 & .65 & .71 & .76 & .84 & .74\\
        \multirow{-2}{*}{WFBE$_{0}$} & 3 & 3 & .65 & .65 & .61 & .65 & .64\\
        \hline
        \hline
        & 1 & 1 & .67 & .79 & .83 & .91 & .80\\
        \multirow{-2}{*}{WFBE$_\text{r}$} & 3 & 1 & .67 & .60 & .63 & .65 & .64\\
        \hline
        & 1 & 3 & .67 & \textbf{.76} & \textbf{.81} & \textbf{.87} & \textbf{.78}\\
        \multirow{-2}{*}{WFBE$_\text{r}$} & 3 & 3 & .67 & \textbf{.70} & \textbf{.65} & \textbf{.68} & \textbf{.68}\\
        \hline
        & 1 & 5 & .67 & .76 & .81 & .88 & .78\\
        \multirow{-2}{*}{WFBE$_\text{r}$} & 3 & 5 & .67 & .70 & .70 & .71 & .70\\
    \end{tabular}
    \caption{MORP's complexity analysis as measured through the ESPL performance metric of heuristic agents: We measure the effect of number of objects $n_o$, number of receptacles $n_r$ and agent capacity $c$ on MORP. For all agents, other metrics were $MC=0.83$, $ES=1$ and $SOR=1$ on average (detailed metrics are shown in Appendix~\ref{ap:heuristics}). Bold numbers indicate best performance in a particular MORP configuration.}
    \vspace{-10pt}
    \label{tab:results}
\end{table}

\subsection{MORP Complexity Analysis}  
In order to analyze the effect of various factors such as agent carrying capacity $c$, total number of objects $n_o$, number of receptacles $n_r$, and navigable area of the scene etc., we consider different configurations of MORP and show the performance of the heuristic and oracle agents on these configurations. Specifically, we vary these factors over a range to create different configurations of MORP:~$c \in {\{1,3\}}$, $n_o \in {\{1,3,5,10\}}$, $n_r \in \{1,2,3\}$. We present the empirical results in Table \ref{tab:results} with these different configurations. The ``medium" and ``large" dataset splits as described in Table~\ref{tab:scenestats} were used for these experiments. Note that we investigate both WFBE$_\text{r}$ and WFBE$_\text{w}$ with $w =\{ 0, 0.5, 1\}$, but we found WFBE$_\text{r}$ to work the best. We therefore use WFBE$_\text{r}$ for most of our conclusions. Next, we summarize our main findings on the effect of the various factors on MORP.\vspace{1pt}

\begin{figure}[h]
    \centering
    \includegraphics[width = 0.5\linewidth]{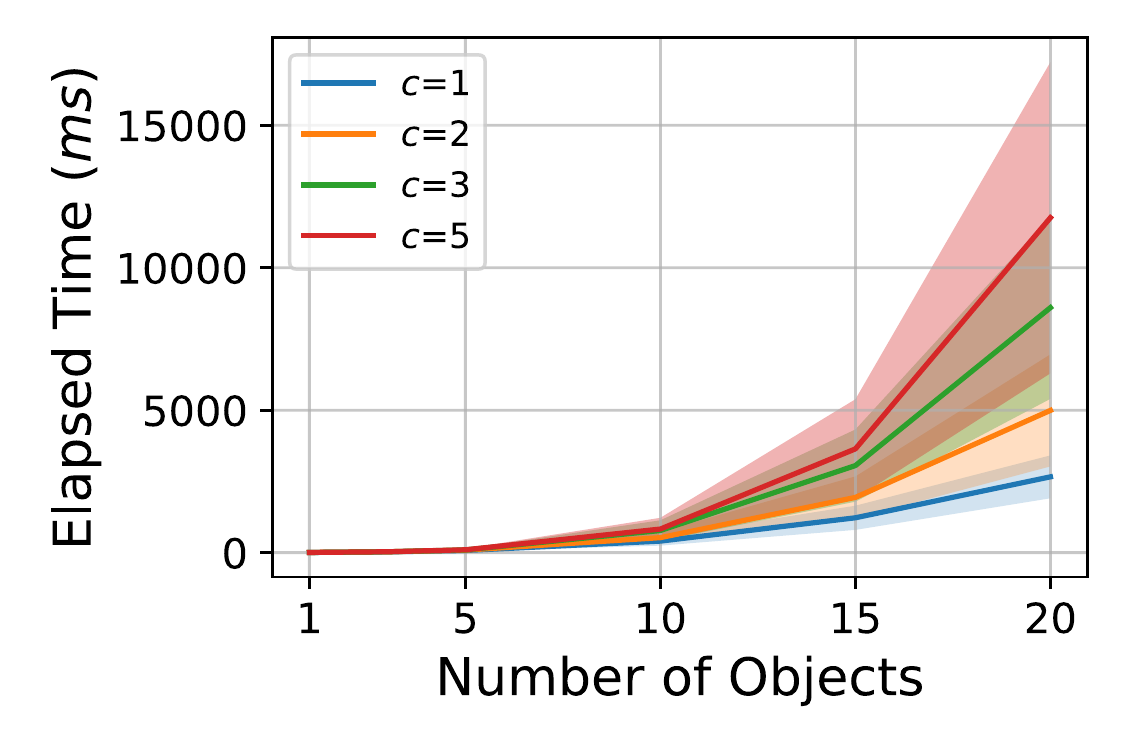}
    \caption{Scalability of oracle agent: time spent by CVRP solver for different values of $c$ and $n_o$.}
    \label{fig:plannertime}
\end{figure}

\noindent\textbf{Higher agent capacity $c$ makes MORP more challenging.} In Table \ref{tab:results}, we observe that all agents' performance decreases with the increase in agent capacity $c$. For instance, when the agent capacity increases from 1 to 3, on average ESPL drops by 10\%. Such increase in MORP's complexity can be attributed to planning. Specifically, fig.~\ref{fig:plannertime} shows increase in planning time taken by the CVRP solver for oracle agents with higher carrying capacity. Although we perform experiments with static agent capacity, practical instantiations of MORP such as house-cleaning might require dynamic capacity through use of containers etc., which may further increase the complexity of MORP.\vspace{1pt}

\noindent\textbf{Higher number of objects reduces exploration complexity.} When there are more objects in the environment, higher percentage of the object discovery happens while planning. For instance, average percentage of objects discovered during the planning $\{26\%, 43\%, 59\%\}$ increases with the total number of objects $\{3,5,10\}$ respectively. This reduces the burden on exploration for object search. Inspite of increase in planning complexity with higher number of objects (see fig.~\ref{fig:plannertime}), such reduction in exploration complexity also reduces the overall problem complexity thereby leading to improved ESPL (Tab.~\ref{tab:results}). Specifically, when the number of objects increases from 1 to 10, on average ESPL for agent with $c=1$ increases by 20\%. Such performance improvement in ESPL however disappears with increase in agent capacity e.g., no significant correlation between $n_o$ and ESPL is observed for $c=3$. This highlights complex interplay between exploration and planning complexity for MORP.\vspace{1pt}

\noindent\textbf{Higher number of receptacles \boldmath{$n_r > 1$} reduces MORP's complexity only when the agent has higher capacity \boldmath{$c > 1$}.} In Table \ref{tab:results}, we see that increase in $n_r$ has different effects on performance of agents with different $c$. When $n_r$ is increased from 1 to 5, we observe that on average ESPL for $c=1$ decreases by 2\%, whereas ESPL for $c=3$ increases by 6\%. It suggests that for episodes with multiple receptacles ($n_r > 1$), there are scenarios where an object's receptacle is within the close proximity of another object. The agents can exploit such scenarios to reduce navigated distance and thereby ESPL for MORP.

\begin{figure}[h]
    \centering
    \includegraphics[width = \linewidth]{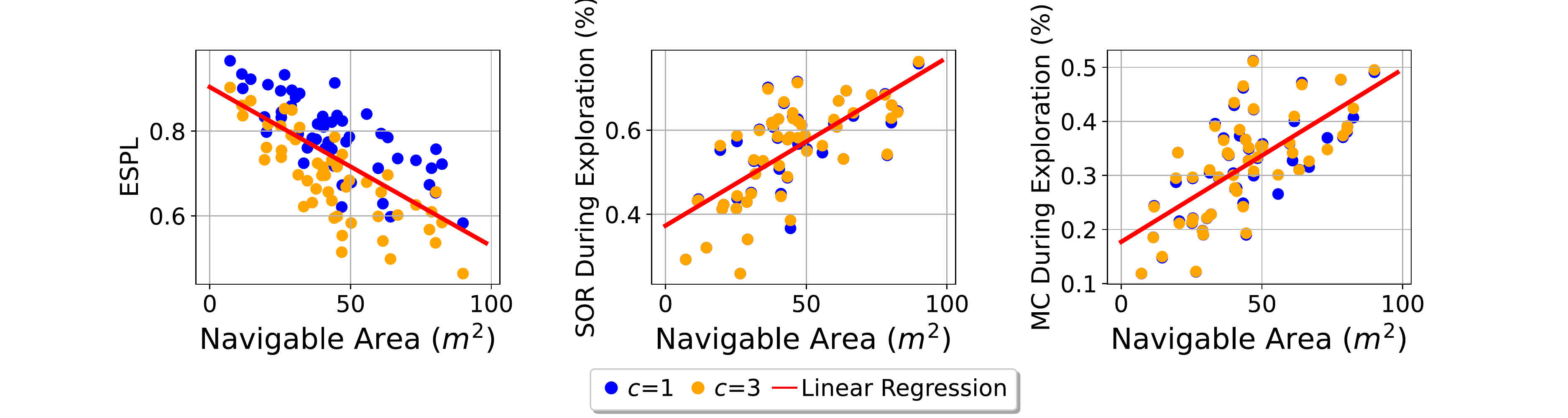}
    \caption{Effect of navigable area on performance as measured by ESPL for WFBE$_\text{r}$ exploration policy.}
    \label{fig:vsarea}
\end{figure}

\noindent\textbf{Greater navigable area worsens the performance.} Figure \ref{fig:vsarea} shows how WFBE$_\text{r}$ exploration policy performance is negatively correlated with navigable area. Similar trends were observed with other exploration strategies. Overall, exploration strategy matters more for larger areas and thus affects performance. We didn't find any correlation between navigation complexity of the scenes (as described in Sec.~\ref{sec:problem}) and performance. 

\subsection{Agent Benchmarking}
We next elaborate on the analysis and failure modes of the classical approaches that we used for exploration and planning in our heuristic and oracle agents. Specifically, we investigate the scaling of the CVRP solver, its applicability to real-time applications, and the performance of different exploration policies.

\subsubsection{Planning} In Figure \ref{fig:plannertime}, we see that the time spent by oracle agent to solve the CVRP problem increases exponentially w.r.t. $n_o$. Figure \ref{fig:plannertime} also shows that increasing $c$ makes the planning more complex. Since our focus is not on solvers, we choose a CVRP solver configuration\footnote{CVRP solver: RoutingModel, FirstSolutionStrategy: PARALLEL-CHEAPEST-INSERTION, solution-limit:50.} from ORTools~\cite{ortools} that can find optimal solution under real-time constraints for our experiments. Based on the exponential increase in compute time for CVRP problems, we recommend using satisficing or learning-based planners e.g.,~\cite{taskogprahy} to tackle larger-scale planning problems in the future.

\subsubsection{Exploration} We find that agents with WFBE$_{r}$ and WFBE$_{50}$ exploration policy outperform all other agents, which suggests that one could potentially learn to find an optimal balance between gain and distance. All other exploration approaches exhibit varied failure modes. For instance, the agents with RND and WFBE$_\text{0}$ exploration policies start picking locations at the opposite sides of the map consecutively and keep going back and forth after exploring for a while. This prevents these agents from exploring unexplored areas. 
Consequently, the agents fail to discover objects and complete the task within maximum episode time ${max}_{t}$. 
Although intuitively choosing the closest frontier location aka WFBE$_{100}$ or frontier with maximum gain i.e., WFBE$_{0}$ perform better than agents with RND exploration policy, they perform worse than the agents that consider both gain and distance aka agents with WFBE$_{r}$ and WFBE$_{50}$. Based on these observations, we also propose WFBE$_{r}$ and its performance on MORP as a baseline for benchmarking future research on MORP. Please see the Appendix for detailed comparison on various exploration approaches using all the metrics defined in Sec.~\ref{sec:problem}. 

\section{Learning-based Agents}

Inspired by the limitations of our heuristic agents and their classical exploration and planning approaches, we explore learning-based agents to obtain policies better than that of the heuristic agents for MORP. In particular, we investigate end-to-end approaches aka monolithic RL for MORP. In addition, we also experiment with ways to improve our heuristic agents. We describe these learning-based agents in detail below. \\




\noindent \textbf{Monolithic agents:} These agents leverage monolithic deep reinforcement learning approach such as~\cite{tabula-rasa} to learn direct mapping from observations to actions.
\begin{itemize}
    \item\textbf{End-to-End Planner (E2E-P)} learns direct mapping from observations to low-level actions -- \emph{forward}, \emph{left}, \emph{right}, \emph{grab/drop}. E2E-P agent thus learns to navigate, explore, and then to accomplish rearrangement from scratch.
    
    \item\textbf{Where-to-Go Planner (W2G-P)} learns where to go on the occupancy map $\mathcal{M}$.
    Instead of low-level navigation actions of \emph{forward}, \emph{left}, \emph{right}, W2G-P agent uses navigable cells from $\mathcal{M}$ as actions in combination with the \emph{grab/drop} action. Once target cell from $\mathcal{M}$ is chosen as the action, we obtain the low-level navigation actions to follow the shortest path between the agent's current location and the chosen cell as described in Sec.~\ref{heuristics}. Unlike E2E-P that has to learn low-level navigation in combination with exploration and planning for MORP, the W2G-P focuses on only learning exploration and planning.
\end{itemize}    

\noindent \textbf{Modular agents:} To improve our modular heuristic agents that greedily combine WFBE-based exploration and CVRP-based planning (Sec.~\ref{sec:benchmark}), we investigate ways to~a)~improve exploration performance,~b)~improve the trade-off between exploration and planning beyond greedy, and~c)~jointly improve both.
\begin{itemize}

    \item\textbf{Learnt Explorer (LE)} is focused on improving exploration performance of WFBE methods in our heuristic agents, inspired by the performance difference between these agents in Table~\ref{tab:results}. Specifically, we learn a utility function approximation for frontier selection in WFBE, which trades-off information gain and frontier distance. We train two types of LE agents:~(1)~LE$_{disc}$ that chooses directly among candidate frontiers, in particular, these candidates correspond to the agent's discrete actions and~(2)~{LE}$_w$~\cite{DRL-FBE} that outputs a continues variable $w \in \mathbb{R}:[0,1]$ to be used as a weight on normalized distance and normalized information gain in WFBE$_\text{w}$ exploration policy. See Appendix~\ref{WFBEAppendix} for more details on WFBE$_\text{w}$ and candidate frontier computation. Contrary to WFBE$_\text{w}$, {LE}$_w$ dynamically changes the $w$ at each step rather than using a fixed $w$. We train LE agents explicitly on the exploration task of finding all objects but not rearranging them for the MORP dataset. We then evaluate the LE agents on MORP by combining them with CVRP planner, akin to the heuristic agents. 
    
    \item\textbf{Optimal Balancer (OB)} agent learns to combine WFBE-based exploration and CVRP-based planning optimally, instead of greedily choosing between them as in the heuristic agents. OB agent thus has two corresponding high-level actions:~\emph{explore} and \emph{plan} that it learns to choose from. 
    
    \item\textbf{Balanced Explorer (BE)} learns exploration policy and optimal balance between \emph{exploration} and \emph{planning} jointly. BE agent's discrete actions thus consists of \emph{plan} action, which calls the CVRP planner; and actions that map to candidate frontier locations.
\end{itemize}

For all of the above agents, the input consists of occupancy map, receptacle locations, agent location, total number of objects to be rearranged, and agent's current gripper state as described in agent's sensor suite (see Sec.~\ref{sec:problem}). For modular agents, we additionally use frontier locations as input. More details on the representations of these inputs for individual agents are described in Appendix~\ref{LearningBasedAppx}. For the policy architecture, we use ConvLSTM-like architecture for E2E-P and W2G-P agents and ConvMLP-like architecture\footnote{Since our inputs for OB, LE, and BE contain all the information pertaining to the sufficient state for MORP, recurrent policies are not needed. We verified this empirically by swapping the MLP layer with the recurrent LSTM layer and found no difference in ESPL of these agents.} for OB, LE, and BE agents. The reward function for all agents is the weighted sum of navigated distance, the number of newly seen objects, grab/drop reward, the newly discovered map area at time step $t$, and the episode success reward. Please see the Appendix~\ref{LearningBasedAppx} for more details.

\subsection{Training Process}
\label{subsec:training}
We use Rllib~\cite{rllib} implementation of DD-PPO~\cite{DDPPO} in order to train our agents with train splits of the dataset. 
E2E-P and W2G-P agents are trained with a naive curriculum, where we first train with single object rearrangement episodes before training with MORP episodes. The ``small" dataset split was used to train E2E-P and W2G-P. All other agents were trained using the ``medium" and ``large" dataset splits (see Table~\ref{tab:scenestats}). Details on task configuration including hyperparameters for training and benchmarking can be found in Appendix~\ref{HyperParameters}.

\subsection{Benchmarking Learning-based Agents}
We evaluate the performance of learning-based agents and compare them against the best performing heuristic agent WFBE$_\text{r}$ in Table~\ref{tab:learntvsheuristic}. We also present our conclusions on leveraging learning to obtain policies for MORP. 

\begin{table}[ht!]
    \fontsize{9pt}{10pt} \selectfont
    \centering
    \rowcolors{1}{gray!3}{gray!7}
    \begin{tabular}{c| c | cccc || cc}
        \multicolumn{1}{c}{} & & 
        \multicolumn{4}{c||}{$ESPL$} & \multicolumn{2}{c}{$\Delta ESPL$}\\
        \hline
        \multicolumn{1}{c|}{Methods} & \diagbox[innerwidth=0.5cm]{$c$}{$n_o$} & \multicolumn{1}{c|}{$1$} & \multicolumn{1}{c|}{$3$} & \multicolumn{1}{c|}{$5$} & \multicolumn{1}{c||}{$10$} & \multicolumn{1}{c|}{$\mu$} & \multicolumn{1}{c}{$\sigma$}\\
        \hline
        & 1  & .67 & \textbf{.76} & \textbf{.81} & \textbf{.87} & - & -\\
        \multirow{-2}{*}{WFBE$_\text{r}$} & 3 & .67 & \textbf{.70} & \textbf{.65} & \textbf{.68} & - & -\\
        \hline
        
        & 1  & \textbf{.68} & .72 & .75 & .85 & .029 & .15\\
        \multirow{-2}{*}{LE$_\text{w}$} & 3  & \textbf{.68} & \textbf{.70} & \textbf{.65} & \textbf{.68} & .002 & .15\\
        \hline
        & 1  & .64 & .69 & .72 & .82 & .062 & .19\\
        \multirow{-2}{*}{LE$_\text{disc}$} & 3  & .64 & .67 & .62 & .65 & .033 & .20\\
        \hline
        & 1 & \textbf{.68} & \textbf{.76} & \textbf{.81} & \textbf{.87} & .007 & .05\\
        \multirow{-2}{*}{OB} & 3 & \textbf{.68} & \textbf{.70} & \textbf{.65} & \textbf{.68} & .006 & .05\\
        \hline
        & 1 & .58 & .65 & .72 & .77 & .097 & .19\\
        \multirow{-2}{*}{BE} & 3 &  .58 & .60 & .58 & .58 & .090 & .18\\
    \end{tabular}
    \caption{Comparison of heuristic and learning-based agents. $\Delta ESPL$ shows mean and variance for $ESPL$ difference between best performing heuristic agent (WFBE$_\text{r}$) and other agents. For all agents, other metrics on average are: $MC=0.83$, $ES=1$ and $SOR=1$. Bold numbers indicate best performance in a particular MORP configuration.}
    \label{tab:learntvsheuristic}
\end{table}

\noindent\textbf{Learning rearrangement planning from scratch is hard.} We find that both E2E-P and W2G-P fail in completing even single object rearrangement episodes. They learn to discover the object and can move towards the object, yet they fail to grab the object or to rearrange it into the receptacle. DRL approaches are known to struggle in long-horizon and sparse reward settings~\cite{Guillaume2020FailureModesRL} as is the case with E2E-P. Likewise, DRL approaches struggle with large discrete actions spaces~\cite{arnold2015LargeActionSpace} such as that of W2G-P. Prior work on object rearrangement therefore uses modular approaches over monolithic ones~\cite{tabula-rasa, szot2021habitat}. We do not include E2E-P and W2G-P metrics in Table~\ref{tab:learntvsheuristic} since these agents do not succeed in any MORP episodes.\vspace{1pt}

\noindent\textbf{Learning exploration explicitly or jointly with rearrangement planning does not help MORP.} We first compare agents that explicitly learn to explore (LE$_\text{w}$ and LE$_\text{disc}$) with WFBE-based exploration on purely exploration task of finding all objects in MORP scenes (see Table~\ref{tab:exploreonlyresults} in Appendix~\ref{LearningBasedAppx} for detailed comparison of the same). We find that learnt exploration agents do not perform better than the WFBE policies when evaluated on this exploration task. Agents with short-sighted frontier-based exploration policies e.g.,  WFBE$_r$ and WFBE$_w$ agents, which try to find objects by exploring most area while navigating minimum distance, perform better than the learnt exploration policies that consider the long-term and minimize total path length in order to discover all the objects in LE$_\text{w}$ and LE$_\text{disc}$ agents. Consequently, combining them with optimal planning using the CVRP solver for rearrangement (LE$_\text{w}$, LE$_\text{disc}$ in Table~\ref{tab:learntvsheuristic}) does not lead to improvement in MORP. This suggests that on average, finding the first object fast and counting on the object discovery during rearrangement planning is a good strategy for MORP. Since LE agents explicitly learn to explore without accounting for rearrangement planning, we also train the BE agent which learns to combines exploration with CVRP planning, while jointly learning to explore. However, we find that this doesn't improve performance on MORP either.\vspace{1pt}

\noindent\textbf{The \add{\textit{greedy combination}} of frontier-based exploration and CVRP-based planning is \add{\textit{empirically}} optimal.} During the training of OB agent, we intermittently evaluate the agent on the test dataset in order to understand the evolution of agent behavior over the training process. Our evaluations indicate that in the early stages of the training, OB acts non-greedily as opposed to the heuristic WFBE$_\text{r}$ agent. Yet, despite hyper-parameter tuning such as that of the entropy loss coefficient in DD-PPO and reward shaping, OB eventually converges to a greedy behaviour. This is evident in $\Delta ESPL$ between OB's performance and WFBE$_\text{r}$ agent's performance in Table \ref{tab:learntvsheuristic}, which is similar to a zero-centered normal distribution with 5\% standard error. Post convergence, OB's high entropy action policies have occasional attempts (10\% of the test data) to explore non-greedily i.e., explore even when there are discovered objects that need to be rearranged. 
Such non-greedy behavior enables OB to discover objects earlier than the  WFBE$_\text{r}$ agent in certain episodes. However, on average OB's behavior is greedy. This empirically demonstrates that the greedy combination of frontier-based exploration and CVRP-based planning (as in WFBE agents) is optimal for MORP.\vspace{1pt}


In summary, E2E-P and W2G-P demonstrate that conventional, monolithic deep RL does not succeed at MORP since the agent needs to learn navigation, exploration, and planning from scratch. Modular agents that leverage the inductive biases of conventional frontier-based exploration and CVRP-based planning in combination with learning to improve exploration and/or the trade-off between exploration and planning still do not outperform their greedy, heuristic counterparts.


\section{Conclusion}
We propose Multiple Object Rearrangement Planning (MORP) in mapped environments with partial observability over object locations as a benchmark task. We conduct a comprehensive complexity analysis of MORP, where we investigate various factors such as number of objects and receptacles, agent carrying capacity, environment layouts etc. that make MORP challenging. We further introduce competitive heuristic baselines for MORP that greedily combine classical frontier-based exploration and optimization-based planning. We also train reinforcement learning policies for MORP in order to outperform the heuristic baselines. However, we find that monolithic RL policies struggle at MORP while modular RL policies converge to behaviors similar to heuristic policies. This empirically demonstrates that greedy combination of exploration and planning for MORP is optimal when objects to be rearranged are uniformly distributed in the 3D environments. However, developing agents that outperform heuristic agents at MORP remains an open problem for future research.


\section*{Acknowledgements}

We'd like to thank Kevin Carlberg and Roberto Calandra for many helpful discussions on this problem.

\bibliography{aaai23}

\begin{thebibliography}{35}
\providecommand{\natexlab}[1]{#1}

\bibitem[{Agia et~al.(2022)Agia, Jatavallabhula, Khodeir, Miksik, Vineet,
  Mukadam, Paull, and Shkurti}]{taskogprahy}
Agia, C.; Jatavallabhula, K.~M.; Khodeir, M.; Miksik, O.; Vineet, V.; Mukadam,
  M.; Paull, L.; and Shkurti, F. 2022.
\newblock Taskography: Evaluating robot task planning over large 3D scene
  graphs.
\newblock In Faust, A.; Hsu, D.; and Neumann, G., eds., \emph{Proceedings of
  the 5th Conference on Robot Learning}, volume 164 of \emph{Proceedings of
  Machine Learning Research}, 46--58. PMLR.

\bibitem[{Anderson et~al.(2018)Anderson, Chang, Chaplot, Dosovitskiy, Gupta,
  Koltun, Kosecka, Malik, Mottaghi, Savva, and Zamir}]{metrics}
Anderson, P.; Chang, A.~X.; Chaplot, D.~S.; Dosovitskiy, A.; Gupta, S.; Koltun,
  V.; Kosecka, J.; Malik, J.; Mottaghi, R.; Savva, M.; and Zamir, A.~R. 2018.
\newblock On Evaluation of Embodied Navigation Agents.
\newblock \emph{CoRR}, abs/1807.06757.

\bibitem[{Andreas, Klein, and Levine(2016)}]{policyskectes}
Andreas, J.; Klein, D.; and Levine, S. 2016.
\newblock Modular Multitask Reinforcement Learning with Policy Sketches.

\bibitem[{Batra et~al.(2020)Batra, Chang, Chernova, Davison, Deng, Koltun,
  Levine, Malik, Mordatch, Mottaghi, Savva, and Su}]{RearrangementChallange}
Batra, D.; Chang, A.~X.; Chernova, S.; Davison, A.~J.; Deng, J.; Koltun, V.;
  Levine, S.; Malik, J.; Mordatch, I.; Mottaghi, R.; Savva, M.; and Su, H.
  2020.
\newblock Rearrangement: {A} Challenge for Embodied {AI}.
\newblock \emph{CoRR}, abs/2011.01975.

\bibitem[{Ben-Shahar and Rivlin(1998)}]{ben1998practical}
Ben-Shahar, O.; and Rivlin, E. 1998.
\newblock Practical pushing planning for rearrangement tasks.
\newblock \emph{IEEE Transactions on Robotics and Automation}, 14(4): 549--565.

\bibitem[{Chaplot et~al.(2020)Chaplot, Gandhi, Gupta, Gupta, and
  Salakhutdinov}]{ActiveSLAM}
Chaplot, D.~S.; Gandhi, D.; Gupta, S.; Gupta, A.; and Salakhutdinov, R. 2020.
\newblock Learning to Explore using Active Neural {SLAM}.
\newblock \emph{CoRR}, abs/2004.05155.

\bibitem[{Chen, Gupta, and Gupta(2019)}]{Exploration1}
Chen, T.; Gupta, S.; and Gupta, A. 2019.
\newblock Learning Exploration Policies for Navigation.
\newblock In \emph{7th International Conference on Learning Representations,
  {ICLR} 2019, New Orleans, LA, USA, May 6-9, 2019}. OpenReview.net.

\bibitem[{Dulac{-}Arnold et~al.(2015)Dulac{-}Arnold, Evans, Sunehag, and
  Coppin}]{arnold2015LargeActionSpace}
Dulac{-}Arnold, G.; Evans, R.; Sunehag, P.; and Coppin, B. 2015.
\newblock Reinforcement Learning in Large Discrete Action Spaces.
\newblock \emph{CoRR}, abs/1512.07679.

\bibitem[{Fukazawa et~al.(2005)Fukazawa, Chomchana, Ota, Yuasa, Arai, Asama,
  and Kawabata}]{JapanRearrangement}
Fukazawa, Y.; Chomchana, T.; Ota, J.; Yuasa, H.; Arai, T.; Asama, H.; and
  Kawabata, K. 2005.
\newblock Realizing the exploration and rearrangement of multiple unknown
  objects by an actual mobile robot.
\newblock \emph{Advanced Robotics}, 19(1): 1--20.

\bibitem[{Garrett et~al.(2021)Garrett, Chitnis, Holladay, Kim, Silver,
  Kaelbling, and Lozano-P{\'e}rez}]{garrett2021integrated}
Garrett, C.~R.; Chitnis, R.; Holladay, R.; Kim, B.; Silver, T.; Kaelbling,
  L.~P.; and Lozano-P{\'e}rez, T. 2021.
\newblock Integrated task and motion planning.
\newblock \emph{Annual review of control, robotics, and autonomous systems}, 4:
  265--293.

\bibitem[{Hall et~al.(2020)Hall, Talbot, Bista, Zhang, Smith, Dayoub, and
  S{\"{u}}nderhauf}]{sceneChangeChallenge}
Hall, D.; Talbot, B.; Bista, S.~R.; Zhang, H.; Smith, R.; Dayoub, F.; and
  S{\"{u}}nderhauf, N. 2020.
\newblock The Robotic Vision Scene Understanding Challenge.
\newblock \emph{CoRR}, abs/2009.05246.

\bibitem[{Hart, Nilsson, and Raphael(1968)}]{A*}
Hart, P.; Nilsson, N.; and Raphael, B. 1968.
\newblock A Formal Basis for the Heuristic Determination of Minimum Cost Paths.
\newblock \emph{{IEEE} Transactions on Systems Science and Cybernetics}, 4(2):
  100--107.

\bibitem[{Huang, Jia, and Mason(2019)}]{large-scale}
Huang, E.; Jia, Z.; and Mason, M.~T. 2019.
\newblock Large-Scale Multi-Object Rearrangement.
\newblock In \emph{2019 International Conference on Robotics and Automation
  (ICRA)}, 211--218.

\bibitem[{Kant et~al.(2022)Kant, Ramachandran, Yenamandra, Gilitschenski,
  Batra, Szot, and Agrawal}]{housekeep}
Kant, Y.; Ramachandran, A.; Yenamandra, S.; Gilitschenski, I.; Batra, D.; Szot,
  A.; and Agrawal, H. 2022.
\newblock Housekeep: Tidying Virtual Households using Commonsense Reasoning.
\newblock arXiv:2205.10712.

\bibitem[{Kapelyukh and Johns(2022)}]{kapelyukh2022my}
Kapelyukh, I.; and Johns, E. 2022.
\newblock My house, my rules: Learning tidying preferences with graph neural
  networks.
\newblock In \emph{Conference on Robot Learning}, 740--749. PMLR.

\bibitem[{Karkus et~al.(2020)Karkus, Mirza, Guez, Jaegle, Lillicrap, Buesing,
  Heess, and Weber}]{tabula-rasa}
Karkus, P.; Mirza, M.; Guez, A.; Jaegle, A.; Lillicrap, T.; Buesing, L.; Heess,
  N.; and Weber, T. 2020.
\newblock Beyond Tabula-Rasa: a Modular Reinforcement Learning Approach for
  Physically Embedded 3D Sokoban.

\bibitem[{Kim et~al.(2021)Kim, Kim, Park, Choi, and Kim}]{Kim2021LearningTE}
Kim, U.-H.; Kim, Y.-H.; Park, J.-M.; Choi, H.-S.; and Kim, J.-H. 2021.
\newblock Learning to Explore, Navigate and Interact for Visual Room
  Rearrangement.

\bibitem[{King, Cognetti, and Srinivasa(2016)}]{objcen-robcen}
King, J.~E.; Cognetti, M.; and Srinivasa, S.~S. 2016.
\newblock Rearrangement planning using object-centric and robot-centric action
  spaces.
\newblock In \emph{2016 IEEE International Conference on Robotics and
  Automation (ICRA)}, 3940--3947.

\bibitem[{King, Ranganeni, and Srinivasa(2017)}]{king2017unobservable}
King, J.~E.; Ranganeni, V.; and Srinivasa, S.~S. 2017.
\newblock Unobservable monte carlo planning for nonprehensile rearrangement
  tasks.
\newblock In \emph{2017 IEEE International Conference on Robotics and
  Automation (ICRA)}, 4681--4688. IEEE.

\bibitem[{Liang et~al.(2018)Liang, Liaw, Nishihara, Moritz, Fox, Goldberg,
  Gonzalez, Jordan, and Stoica}]{rllib}
Liang, E.; Liaw, R.; Nishihara, R.; Moritz, P.; Fox, R.; Goldberg, K.;
  Gonzalez, J.~E.; Jordan, M.~I.; and Stoica, I. 2018.
\newblock {RLlib}: Abstractions for Distributed Reinforcement Learning.
\newblock In \emph{International Conference on Machine Learning ({ICML})}.

\bibitem[{Liu et~al.(2018)Liu, Lehman, Molino, Such, Frank, Sergeev, and
  Yosinski}]{liu2018CoordConv}
Liu, R.; Lehman, J.; Molino, P.; Such, F.~P.; Frank, E.; Sergeev, A.; and
  Yosinski, J. 2018.
\newblock An Intriguing Failing of Convolutional Neural Networks and the
  CoordConv Solution.
\newblock \emph{CoRR}, abs/1807.03247.

\bibitem[{Lloyd(1982)}]{Kmeans}
Lloyd, S. 1982.
\newblock Least squares quantization in PCM.
\newblock \emph{IEEE Transactions on Information Theory}, 28(2): 129--137.

\bibitem[{Matheron, Perrin, and Sigaud(2020)}]{Guillaume2020FailureModesRL}
Matheron, G.; Perrin, N.; and Sigaud, O. 2020.
\newblock Understanding Failures of Deterministic Actor-Critic with Continuous
  Action Spaces and Sparse Rewards.
\newblock In Farka{\v{s}}, I.; Masulli, P.; and Wermter, S., eds.,
  \emph{Artificial Neural Networks and Machine Learning -- ICANN 2020},
  308--320. Cham: Springer International Publishing.
\newblock ISBN 978-3-030-61616-8.

\bibitem[{Newman, Carlberg, and Desai(2020)}]{kondo}
Newman, B.; Carlberg, K.; and Desai, R. 2020.
\newblock Optimal Assistance for Object-Rearrangement Tasks in Augmented
  Reality.
\newblock \emph{CoRR}, abs/2010.07358.

\bibitem[{Niroui et~al.(2019)Niroui, Zhang, Kashino, and Nejat}]{DRL-FBE}
Niroui, F.; Zhang, K.; Kashino, Z.; and Nejat, G. 2019.
\newblock Deep Reinforcement Learning Robot for Search and Rescue Applications:
  Exploration in Unknown Cluttered Environments.
\newblock \emph{IEEE Robotics and Automation Letters}, 4(2): 610--617.

\bibitem[{Perron and Furnon(2022)}]{ortools}
Perron, L.; and Furnon, V. 2022.
\newblock OR-Tools.
\newblock Google.

\bibitem[{Ramakrishnan et~al.(2021)Ramakrishnan, Gokaslan, Wijmans, Maksymets,
  Clegg, Turner, Undersander, Galuba, Westbury, Chang, Savva, Zhao, and
  Batra}]{hm3d}
Ramakrishnan, S.~K.; Gokaslan, A.; Wijmans, E.; Maksymets, O.; Clegg, A.;
  Turner, J.~M.; Undersander, E.; Galuba, W.; Westbury, A.; Chang, A.~X.;
  Savva, M.; Zhao, Y.; and Batra, D. 2021.
\newblock Habitat-Matterport 3D Dataset ({HM}3D): 1000 Large-scale 3D
  Environments for Embodied {AI}.
\newblock In \emph{Thirty-fifth Conference on Neural Information Processing
  Systems Datasets and Benchmarks Track (Round 2)}.

\bibitem[{Ramakrishnan, Jayaraman, and Grauman(2020)}]{EforEVE}
Ramakrishnan, S.~K.; Jayaraman, D.; and Grauman, K. 2020.
\newblock An Exploration of Embodied Visual Exploration.
\newblock \emph{CoRR}, abs/2001.02192.

\bibitem[{Ronneberger, Fischer, and Brox(2015)}]{ronneberger2015UNet}
Ronneberger, O.; Fischer, P.; and Brox, T. 2015.
\newblock U-Net: Convolutional Networks for Biomedical Image Segmentation.
\newblock \emph{CoRR}, abs/1505.04597.

\bibitem[{Szot et~al.(2021)Szot, Clegg, Undersander, Wijmans, Zhao, Turner,
  Maestre, Mukadam, Chaplot, Maksymets, Gokaslan, Vondrus, Dharur, Meier,
  Galuba, Chang, Kira, Koltun, Malik, Savva, and Batra}]{szot2021habitat}
Szot, A.; Clegg, A.; Undersander, E.; Wijmans, E.; Zhao, Y.; Turner, J.;
  Maestre, N.; Mukadam, M.; Chaplot, D.; Maksymets, O.; Gokaslan, A.; Vondrus,
  V.; Dharur, S.; Meier, F.; Galuba, W.; Chang, A.; Kira, Z.; Koltun, V.;
  Malik, J.; Savva, M.; and Batra, D. 2021.
\newblock Habitat 2.0: Training Home Assistants to Rearrange their Habitat.
\newblock In \emph{Advances in Neural Information Processing Systems
  (NeurIPS)}.

\bibitem[{Weihs et~al.(2021)Weihs, Deitke, Kembhavi, and Mottaghi}]{VisualRe}
Weihs, L.; Deitke, M.; Kembhavi, A.; and Mottaghi, R. 2021.
\newblock Visual Room Rearrangement.
\newblock In \emph{{IEEE} Conference on Computer Vision and Pattern
  Recognition, {CVPR} 2021, virtual, June 19-25, 2021}, 5922--5931. Computer
  Vision Foundation / {IEEE}.

\bibitem[{Wijmans et~al.(2020)Wijmans, Kadian, Morcos, Lee, Essa, Parikh,
  Savva, and Batra}]{DDPPO}
Wijmans, E.; Kadian, A.; Morcos, A.; Lee, S.; Essa, I.; Parikh, D.; Savva, M.;
  and Batra, D. 2020.
\newblock {DD-PPO:} Learning Near-Perfect PointGoal Navigators from 2.5 Billion
  Frames.
\newblock In \emph{8th International Conference on Learning Representations,
  {ICLR} 2020, Addis Ababa, Ethiopia, April 26-30, 2020}. OpenReview.net.

\bibitem[{Xia et~al.(2018)Xia, R.~Zamir, He, Sax, Malik, and
  Savarese}]{gibsonenv}
Xia, F.; R.~Zamir, A.; He, Z.-Y.; Sax, A.; Malik, J.; and Savarese, S. 2018.
\newblock Gibson env: real-world perception for embodied agents.
\newblock In \emph{Computer Vision and Pattern Recognition (CVPR), 2018 IEEE
  Conference on}. IEEE.

\bibitem[{Yamauchi(1997)}]{FBE}
Yamauchi, B. 1997.
\newblock A frontier-based approach for autonomous exploration.
\newblock In \emph{Proceedings 1997 IEEE International Symposium on
  Computational Intelligence in Robotics and Automation CIRA'97. 'Towards New
  Computational Principles for Robotics and Automation'}, 146--151.

\bibitem[{Yamauchi(1998)}]{yamauchi1998frontier}
Yamauchi, B. 1998.
\newblock Frontier-based exploration using multiple robots.
\newblock In \emph{Proceedings of the second international conference on
  Autonomous agents}, 47--53.

\end{thebibliography}

\clearpage

\appendix
\section{Appendix}
\subsection{Dataset}

\begin{figure*}[bht!]
    \centering
    \includegraphics[width = \linewidth]{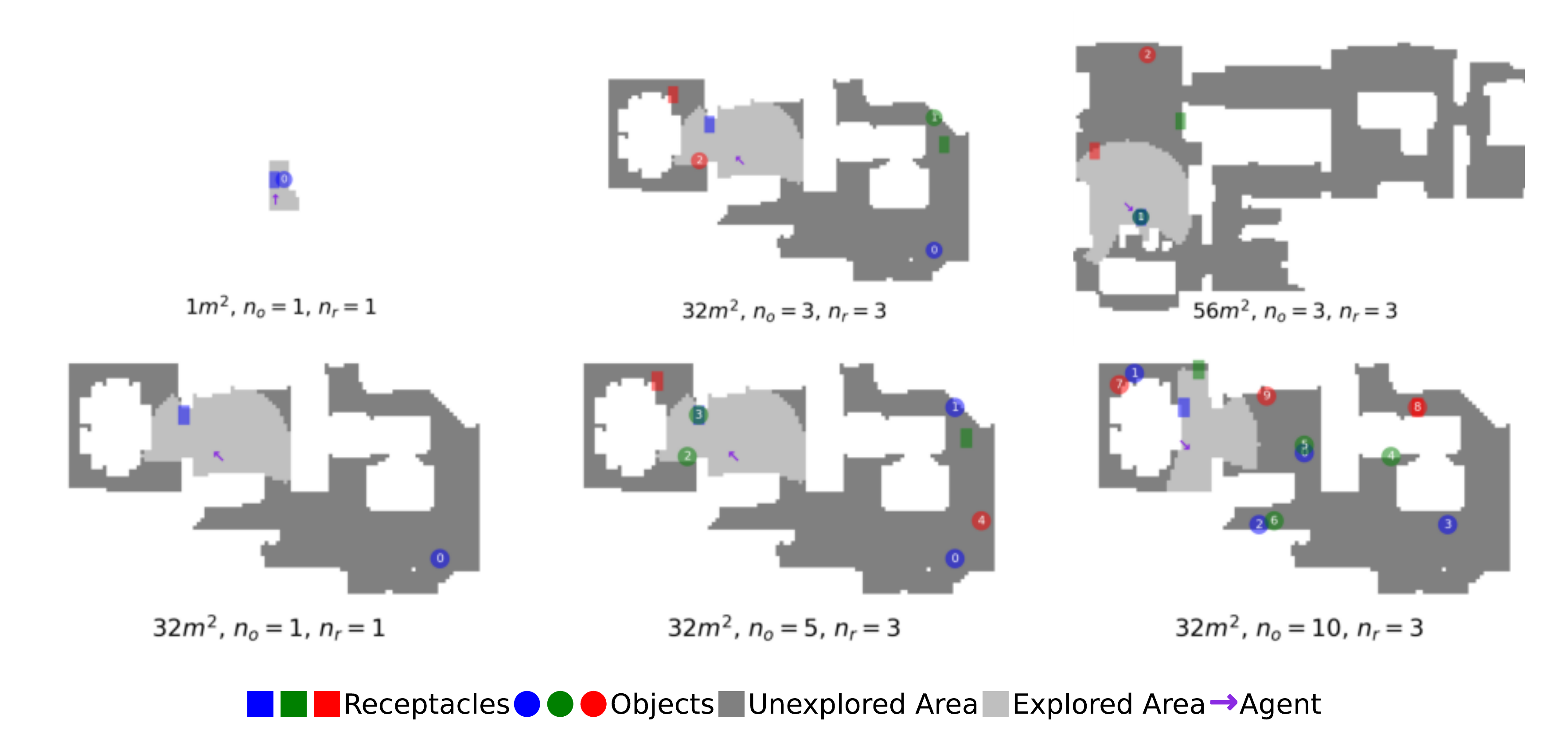}
    \caption{Instantiation of multiple episodes. The top row shows different scenes that belong to small/medium/large splits of our dataset. The bottom row shows the instantiation of a single scene with varying number of objects $n_o$ and receptacles $n_r$.}
    \label{fig:canonicalscene}
\end{figure*}

Table~\ref{tab:episodestats} shows the number of scenes and episodes for the small, medium and large variants of our dataset. Figure~\ref{fig:canonicalscene} visualizes multiple episodes in our dataset variants with varying number of objects and receptacles.

\begin{table}[h!]
    \rowcolors{1}{gray!3}{gray!7}
    \centering
    \begin{tabular}{c c|c | cccc}
        & & \#scenes & \multicolumn{4}{c}{\#episodes}\\
        \hline
        & $n_o$ & - & 1 & 3 & 5 & 10\\
        \hline
        large & train & 44 & 4.4k & 4.4k & 4.4k & 4.4k \\
        & test & 22 & 220 & 220 & 220 & 220\\
        \hline
        medium & train & 60 & 6k & 6k & 6k & 4843\\
        & test & 30 & 300 & 300 & 299 & 231\\
        \hline
        small & train & 5 & 500 & - & - & -\\
        & test & 1 & 10 & - & - & - \\
    \end{tabular}
    \caption{Episode and scene counts in the dataset.}
    \label{tab:episodestats}
\end{table}

\subsection{Capacitated Vehicle Routing Problem}
\label{CVRPFormulation}
In this section, we describe the CVRP formulation employed by the oracle agent. Let $\mathbb{O}=\{0, ..., n_s-1\}$ denote the set of $n_s$ objects seen and yet to be rearranged. Then the set of object pickup locations and their corresponding receptacle locations yields $2n_s+1$ locations in total including the agent's initial location $x$. We order these locations with indices $\{0, ..., 2n_s\}$ such that the zeroth index corresponds to $x$, every odd index $2i+1$ corresponds to the pickup location in $\mathcal{P}$ and every even index $2i+2$ corresponds to the dropoff location in $\mathcal{B}$ for object $i \in \mathbb{O}$. We denote the set of pickup locations as $\Gamma = \{1, 3, ..., 2n_s-1\}$ and the set of dropoff locations as $\Psi=\{2, 4, ..., 2n_s\}$.

Let $\tau: \{0, ..., 2n_s\} \times \{0, ..., 2n_s\} \rightarrow{\mathbb{R}}$ be the transportation cost function which calculates the geodesic distance between two navigable locations given the location indices and satisfies $\tau(i,i)=0$.
We define $\varphi: \{0...2n_s\} \rightarrow \{0...2n_s\}$ as an invertible operator that maps a step number to a location index to visit. The operator $\varphi$ is the solution variable for our oracle planner, so the optimal sequence of location visits is given by $({\varphi^*(0)},...,\varphi^*(2n_s))$ and is computed as the solution to a combinatorial optimization problem. Invertibility is assumed for $\varphi$ since every location is visited exactly once and each visit (except the zeroth location $x$) is implicitly tied to either an object pickup or dropoff. Let $\mathbf{1_I(\cdot)}$ be the indicator function to check if its argument is in the set $\mathbf{I}$. The CVRP optimization problem is then framed as:

\begin{subequations}\label{cvrp:main}
    \fontsize{9pt}{10pt} \selectfont
    \begin{align}
        \underset{(\varphi(0),...\varphi(2n_s))}{\min}&\sum_{i=1}^{2n_s}{\tau(\varphi(i), \varphi(i-1))} \label{cvrp:obj}\\
        \text{s.t. } &\varphi(0) = 0 \label{cvrp:visitx}\\
        &\sum_{i=1}^{j}{\mathbf{1}_{\Gamma}(\varphi(i))-\mathbf{1}_{\Psi}(\varphi(i)) \leq c}, \forall j \in {\{1,...,2n_s\}} \label{cvrp:capacity}\\
        &\varphi^{-1}(i+1) > \varphi^{-1}(i), \forall i \in \Gamma \label{cvrp:pickdrop}\\
        &\varphi^{-1}(2i+1) \leq {||\mathcal{H}||}_1, \forall i \in \{\mathbb{O} \hspace{2pt}|\hspace{4pt} {||\mathcal{H}_{:,i}||}_1 = 1\} \label{cvrp:visited}
    \end{align}
\end{subequations}

The objective function in eq~\eqref{cvrp:obj} corresponds to the total distance traveled. Eq~\eqref{cvrp:visitx} states that the initial location of the agent should be the first location visited. Eq~\eqref{cvrp:capacity} expresses the capacity constraint for the agent. Eq~\eqref{cvrp:pickdrop} indicates that the agent should pickup the object before dropping it to its receptacle. Eq~\eqref{cvrp:visited} use the gripper slot representation $\mathcal{H} \in \mathbb{R}^{c \times n_o}$ defined in Section \ref{subsec:agent_def} to maintain consistency for held objects when re-planning occurs and is trivially satisfied if no object is being held. It states that the pickup for held objects must happen before others. But since they are already being held in the environment, no actual re-pickup needs to be performed during plan execution.

\subsection{Weighted Frontier Based Exploration}
\label{WFBEAppendix}
We detail our implementation of WFBE along with the utility function definitions in this section.

In our implementation, we first compute the FBE frontiers and then cluster them into $K$ clusters\footnote{We empirically chose the number of clusters $K$ to be $10$.} using K-means clustering~\cite{Kmeans}. Denoting the set $\{1, \ldots, K\}$ by $[K]$, let $\mathcal{V}$ be the set of $K$ frontier locations with element $v_k \in \mathbb{Z}^2$ being the frontier location closest to the centroid of cluster $k \in [K]$. Define $d_k$ to be the shortest path length between $v_k$ and $x$. Further, let $gain_k \in \mathbb{Z}$ be the gain for the candidate frontier $v_k$ defined as the sum of newly seen cells along the shortest path between $x$ and $v_k$.

We explore two choices to calculate the utility function for choosing the next frontier:
\begin{itemize}
    \item \textbf{WFBE$_\text{r}$}: Chooses the candidate frontier which maximizes the ratio between gain and distance of the frontier from $x$,
    \begin{equation}
        choice = \underset{k \in [K]}{\text{arg}\max}\left\{ \frac{gain_k}{d_k} \right\}
    \end{equation}
    
    \item \textbf{WFBE$_\text{w}$}: Chooses the candidate frontier which minimizes a weighted sum of normalized gain and normalized distance from $x$,
    \begin{equation} \label{eq:wcoeeff}
       choice = \underset{k \in [K]}{\text{arg}\min} \text{ } \{ w*\bar{d}_k + (1-w)*\overline{gain}_k \},
    \end{equation}
    \begin{equation*}
        \text{where } \bar{d}_k = \frac{d_k}{\sum_{j \in [K]} d_j}, \overline{gain}_k = \frac{gain_k}{\sum_{j \in [K]} gain_j},
    \end{equation*}
     and $w$ is a balancing hyperparameter~\cite{DRL-FBE}.
\end{itemize}

\subsection{Complexity analysis of MORP}
\label{ap:heuristics}
Table~\ref{tab:fullresults} shows the performance metrics for heuristic agents when the number of objects $n_o$, number of receptacles $n_r$ and the agent capacity $c$ vary.

\begin{table*}[h]
    \fontsize{9pt}{10pt} \selectfont
    \centering
    \rowcolors{1}{gray!3}{gray!7}
    \begin{tabular}{c|c|c|cccc|cccc|cccc|cccc||c}
        \multicolumn{1}{c}{ } &\multicolumn{1}{c}{}&\multicolumn{1}{c|}{}&
        \multicolumn{4}{c|}{$MC$}&
        \multicolumn{4}{c|}{$SOR$}&
        \multicolumn{4}{c|}{$ROR$}&
        \multicolumn{5}{c}{$ESPL$}\\
        \hline
         \textit{Explorer} & $c$ & \diagbox[innerwidth=0.5cm]{$n_r$}{$n_o$} &
         \multicolumn{1}{c|}{$1$} & \multicolumn{1}{c|}{$3$} & \multicolumn{1}{c|}{$5$} & \multicolumn{1}{c|}{$10$} &
         \multicolumn{1}{c|}{$1$} & \multicolumn{1}{c|}{$3$} & \multicolumn{1}{c|}{$5$} & \multicolumn{1}{c|}{$10$} &
         \multicolumn{1}{c|}{$1$} & \multicolumn{1}{c|}{$3$} & \multicolumn{1}{c|}{$5$} & \multicolumn{1}{c|}{$10$} &
         \multicolumn{1}{c|}{$1$} & \multicolumn{1}{c|}{$3$} & \multicolumn{1}{c|}{$5$} & \multicolumn{1}{c||}{$10$} & \multicolumn{1}{c}{\textit{avg}}\\
        \hline
        & 1 & 3 & .66 & .84 & .89 & .94 & 1 & 1 & 1 & 1 & 1 & 1 & 1 & 1 & .62 & .69 & .74 & .82 & .72\\
        \multirow{-2}{*}{RND} & 3 & 3 & .66 & .84 & .89 & .94 & 1 & 1 & 1 & 1 & 1 & 1 & 1 & 1 & .62 & .63 & .60 & .62 & .62\\
        \hline
        & 1 & 3 & .66 & .85 & . 89 & .94 & 1 & 1 & 1 & 1 & 1 & 1 & 1 & 1 & .56 & .67 & .75 & .83 & .71\\
        \multirow{-2}{*}{WFBE$_{1}$} & 3 & 3 & .66 & .85 & .87 & .94 & 1 & 1 & 1 & 1 & 1 & 1 & 1 & 1 & .56 & .62 & .61 & .64 & .61\\
        \hline
        & 1 & 3 & .65 & .84 & .88 & .94 & 1 & 1 & 1 & 1 & 1 & 1 & 1 & 1 & \textbf{.68} & .75 & \textbf{.81} & \textbf{.87} & \textbf{.78}\\
        \multirow{-2}{*}{WFBE$_{0.5}$} & 3 & 3 & .65 & .84 & .89 & .94 & 1 & 1 & 1 & 1 & 1 & 1 & 1 & 1 & \textbf{.68} & .69 & \textbf{.65} & .67 & .67\\
        \hline
        & 1 & 3 & .65 & .84 & .89 & .94 & 1 & 1 & 1 & 1 & 1 & 1 & 1 & 1 & .65 & .71 & .76 & .84 & .74\\
        \multirow{-2}{*}{WFBE$_{0}$} & 3 & 3 & .65 & .84 & .89 & .93 & 1 & 1 & 1 & 1 & 1 & 1 & 1 & 1 & .65 & .65 & .61 & .65 & .64\\
        \hline
        & 1 & 3 & .65 & .84 & .89 & .94 & 1 & 1 & 1 & 1 & 1 & 1 & 1 & 1 & .67 & \textbf{.76} & \textbf{.81} & \textbf{.87} & \textbf{.78}\\
        \multirow{-2}{*}{WFBE$_\text{r}$} & 3 & 3 & .65 & .84 & .89 & .93 & 1 & 1 & 1 & 1 & 1 & 1 & 1 & 1 & .67 & \textbf{.70} & \textbf{.65} & \textbf{.68} & \textbf{.68}\\
        \hline
        & 1 & 3 & .65 & .80 & .87 & .94 & 1 & 1 & 1 & 1 & 1 & 1 & 1 & 1 & \textbf{.68} & .72 & .75 & .85 & .75\\
        \multirow{-2}{*}{LE$_\text{w}$} & 3 & 3 & .65 & .80 & .87 & .94 & 1 & 1 & 1 & 1 & 1 & 1 & 1 & 1 & \textbf{.68} & \textbf{.70} & \textbf{.65} & \textbf{.68} & \textbf{.68}\\
        \hline
        & 1 & 3 & .65 & .79 & .87 & .94 & 1 & 1 & 1 & 1 & 1 & 1 & 1 & 1 & .67 & .64 & .69 & .72 & .68\\
        \multirow{-2}{*}{LE$_\text{disc}$} & 3 & 3 & .65 & .79 & .87 & .94 & 1 & 1 & 1 & 1 & 1 & 1 & 1 & 1 & .67 & .64 & .67 & .62 & .65\\
        \hline
        & 1 & 3 & .65 & .84 & .89 & .94 & 1 & 1 & 1 & 1 & 1 & 1 & 1 & 1 & \textbf{.68} & \textbf{.76} & \textbf{.81} & \textbf{.87} & \textbf{.78}\\
        \multirow{-2}{*}{OB} & 3 & 3 & .65 & .84 & .89 & .94 & 1 & 1 & 1 & 1 & 1 & 1 & 1 & 1 & \textbf{.68} & \textbf{.70} & \textbf{.65} & \textbf{.68} & \textbf{.68}\\
        \hline
        & 1 & 3 & .66 & .85 & .90 & .94 & 1 & 1 & 1 & 1 & 1 & 1 & 1 & 1 & .58 & .65 & .82 & .77 & .68\\
        \multirow{-2}{*}{BE} & 3 & 3 & .66 & .85 & .89 & .94 & 1 & 1 & 1 & 1 & 1 & 1 & 1 & 1 & .58 & .60 & .58 & .58 & .59\\
    \end{tabular}
    \caption{MORP's complexity analysis as measured through the ESPL performance metric of heuristic agents: We measure the effect of number of objects $n_o$, number of receptacles $n_r$ and agent capacity $c$ on MORP.}
    \label{tab:fullresults}
\end{table*}

\subsection{Learning-Based Agents Implementation Details}
\label{LearningBasedAppx}
In this section, we describe the policy neural network architectures, reward functions and training hyperparameters for our learning-based agents. 

\subsubsection{Inputs}
We first define a common set of inputs for learning-based agents. The inputs to individual agents are shown in Table~\ref{tab:statespace}.
\begin{itemize}
    \item Agent Location Map: A matrix $\mathcal{M}_a \in \mathbb{Z}^{200 \times 150}$ with all entries 0s, except the agent location which is a 1.
    \item Expected Number of Objects Map: $\mathcal{M}_o: \mathbb{R}^{n_r \times 200 \times 150}$ which contains the expected number of objects of type $i$ in the $i$-th slice $\mathcal{M}_{o,i,:,:}$. The matrix $\mathcal{M}_o$ is initialized with zeros. Every seen object of type $i$ contributes a further additive value of $1$ and unseen type $i$ objects contribute the additive value $\frac{1}{\text{\# of unseen navigable cells}}$ to the $i$-th slice.
    \item Final Object Locations Map: $\mathcal{M}_b: \mathbb{R}^{n_r \times 200 \times 150}$ with the $i$-th slice $\mathcal{M}_{b,i,:,:}$ containing the number of type $i$ objects at the $i$-th receptacle location and zero otherwise.
    \item Candidate Frontier Map: $\mathcal{M}_v \in \mathbb{Z}^{200 \times 150}$ with candidate frontier locations represented as ones and all other locations as zeros.
    \item Candidate Frontier Distances and Gains: $\mathcal{D} = \{d_{v_i}, \forall v_i \in \mathcal{V}\}$ distances and $\mathcal{K} = \{gain_{v_i}, \forall v_i \in \mathcal{V}\}$ gains associated with candidate frontiers.
    \item Exploration Location Map: $M_{e}: \mathbb{Z}^{200 \times 150}$ which represents the selected exploration location with 1 and all other locations with 0.
    \item Planning Location Map: $M_{p}: \mathbb{Z}^{200 \times 150}$ which represents the next planned object/receptacle location as 1 and all other locations as 0.
    \item Gripper Load: $g_l = \frac{||\mathcal{H}||_1}{c}$, continuous representation of gripper load in $[0,1]$.
\end{itemize}

\subsubsection{Policy Architectures}
In order to present policy architectures we define individual and combined layers as follows:
\begin{itemize}
    \item \textit{conv1:} 8 $(4\times4)$ kernels, padding:1, stride:2 $+$ ReLU.
    \item \textit{conv2:} 16 $(4\times4)$ kernels, padding:1, stride:2 $+$ ReLU.
    \item \textit{conv3:} 16 $(3\times3)$ kernels, padding:1, stride:2 $+$ ReLU.
    \item \textit{conv4:} 8 $(3\times3)$ kernels, padding:1, stride:2 $+$ ReLU.
    \item \textit{conv5:} 4 $(3\times3)$ kernels, padding:1, stride:1 $+$ ReLU.
    \item \textit{conv6:} 2 $(2\times2)$ kernels, padding:1, stride:1 $+$ ReLU.
    \item \textit{conv7:} 32 $(3\times3)$ kernels, padding:1, stride:1 $+$ ReLU.
    \item \textit{flat1:} FC layer from 1900-dim to 512-dim $+$ ReLU.
    \item \textit{flat2:} FC layer from 3800-dim to 512-dim $+$ ReLU.
    \item \textit{concat1:} FC layer from 1545-dim to 256-dim $+$ ReLU.
    \item \textit{concat2:} FC layer from 1609-dim to 256-dim $+$ ReLU.
    \item \textit{concat3:} LSTM layer from 1549-dim to 256-dim.
    \item \textit{concat4:} LSTM layer from 1613-dim to 256-dim.
    \item \textit{concat5:} FC layer from 585-dim to 256-dim $+$ ReLU.
    \item \textit{concat6:} LSTM layer from 521-dim to 256-dim.
    \item \textit{act-emb:} Embeds action into 4-dim cont. space.
    \item \textit{dir-emb:} Embeds direction into 8-dim cont. space.
    \item \textit{info:} FC layer from 20-dim to 64-dim $+$ ReLU.
    \item \textit{c1:}$flat1(conv5(conv4(conv4(conv1(\cdot)))))$
    \item \textit{c2:}$flat1(conv5(conv4(conv3(conv2(\cdot)))))$
    \item \textit{c3:}$flat2(conv4(conv3(conv7(conv2(\cdot)))))$
    \item \textit{act:} Action head FC layer from 256-dim to actions.
    \item \textit{cri:} Critic head FC layer from 256-dim to 1-dim.
\end{itemize}

Learning-based agents' policy architectures are as follows:
\begin{itemize}
    \item \textbf{LE$_w$, LE$_{disc}$:} \textit{concat5}([\textit{c3}($\mathcal{M}, \mathcal{M}_a, \mathcal{M}_o, \mathcal{M}_v$), \textit{dir-emb}($\phi$), \textit{info}([$\mathcal{K}, \mathcal{D}$])]).
    \item \textbf{E2E-P:} \textit{concat6}([\textit{c3}($\mathcal{M}, \mathcal{M}_a, \mathcal{M}_o, \mathcal{M}_b$), \textit{dir-emb}($\phi$), $g_l$)]).
    \item \textbf{W2G-P:} We use E2E-P network to obtain state representation and $200 \times 150$ top-down map, action logits, is reconstructed by using \emph{up-sampling} similar to U-Net~\cite{ronneberger2015UNet} combined with \emph{coordinate convolution}~\cite{liu2018CoordConv}.
    \item \textbf{OB(MLP):} \textit{concat1}([\textit{c1}($\mathcal{M}$), \textit{c2}($\mathcal{M}_b, \mathcal{M}_e, \mathcal{M}_p$), \textit{c1}($\mathcal{M}_o, \mathcal{M}_b$), \textit{dir-emb}($\phi$), $g_l$]).
    \item \textbf{OB(LSTM):} \textit{concat3}([\textit{c1}($\mathcal{M}$), \textit{c2}($\mathcal{M}_b, \mathcal{M}_e, \mathcal{M}_p$), \textit{c1}($\mathcal{M}_o, \mathcal{M}_b$), \textit{dir-emb}($\phi$), \textit{act-emb}(\textit{prev-action}), $g_l$]).
    \item \textbf{BE(MLP):} \textit{concat2}([\textit{c1}($\mathcal{M}$), \textit{c2}($\mathcal{M}_b, \mathcal{M}_v, \mathcal{M}_p$), \textit{c1}($\mathcal{M}_o, \mathcal{M}_b$), \textit{dir-emb}($\phi$), $g_l$, \textit{info}([$\mathcal{K}, \mathcal{D}$])]).
    \item \textbf{BE(LSTM):} \textit{concat4}([\textit{c1}($\mathcal{M}$), \textit{c2}($\mathcal{M}_b, \mathcal{M}_v, \mathcal{M}_p$), \textit{c1}($\mathcal{M}_o, \mathcal{M}_b$), \textit{dir-emb}($\phi$), \textit{act-emb}(\textit{prev-action}), $g_l$, \textit{info}([$\mathcal{K}, \mathcal{D}$])]).
\end{itemize}

We feed output of \textit{concatX} layers into \textit{act} and \textit{cri} layers in order to produce the actor and the critic in the DD-PPO algorithm.

\subsubsection{Reward Functions}
The reward function for all agents is the weighted sum of navigated distance, the number of newly seen objects, grab/drop reward, the newly discovered map area at time step $t$, and the episode success reward. Letting $\Delta{x}_t = d(x_t, x_{t-1})$ be the distance navigated by agent at $t$, $\Delta{SO}_t = ({SOR}_t-{SOR}_{t-1})*n_o$ be the change in the number of seen objects at time $t$, $\Delta{RO}_t = ({ROR}_t-{ROR}_{t-1})*n_o$ be the change in the number of rearranged objects at time $t$, $\Delta{SC}_t = ({MC}_t-{MC}_{t-1}) \times \{\text{total number of navigable cells in the map}\}$ be the number of newly seen cells at time $t$, $\Delta{h}_t = c*({g_l}_t - {g_l}_{t-1})$ be the change in the number of held objects at time $t$, and $ES$ be the episode success from Section~\ref{subsec:evaluation}, we can define the reward functions as follows:
\begin{subequations}
    \begin{multline}\label{reward:Modular}
        R_t^{1} = -3\Delta{x}_t + 20(\Delta{RO}_t + \Delta{SO}_t)\\ + 10^{-2}\Delta{SC}_t + 500ES
    \end{multline}
    \begin{multline}\label{reward:Monolithic}
        R_t^2 = -3\Delta{x}_t + 20(\Delta{RO}_t + \Delta{SO}_t\\ + \Delta{h}_t) + 10^{-2}\Delta{SC}_t
    \end{multline}
    \begin{equation}\label{reward:ESPL}
    R_t^3 = 
    \left\{
        \begin{array}{lr}
            0, & t \neq T\\
            e^{ESPL}, & t = T
        \end{array}\right\}
\end{equation}
\end{subequations}
These rewards are then used to train the various learning-based agents as shown in Table~\ref{tab:statespace}.

\begin{table}[h!]
    \rowcolors{1}{gray!3}{gray!7}
    \centering
    \begin{tabular}{c|c|c}
        Agent & Inputs & Reward\\ \hline
        LE$_{disc}$ , LE$_{w}$ & $\mathcal{M}_v, \mathcal{K}, \mathcal{D}$ & Eq~\eqref{reward:Modular}\\
        E2E-P, W2G-P & $\mathcal{M}_{b}, g_l$ & Eq~\eqref{reward:Monolithic}\\
        OB(MLP) & $\mathcal{M}_{b}, \mathcal{M}_{e}, \mathcal{M}_{p}, g_l$ & Eq~\eqref{reward:Modular}\\
        OB(LSTM) & $\mathcal{M}_{b}, \mathcal{M}_{e}, \mathcal{M}_{p}, g_l$ & Eq~\eqref{reward:ESPL}\\
        BE(MLP) & $ \mathcal{M}_{b}, \mathcal{M}_v, \mathcal{M}_{p},\mathcal{K}, \mathcal{D}, g_l$ & Eq~\eqref{reward:Modular}\\
        BE(LSTM) & $ \mathcal{M}_{b}, \mathcal{M}_v, \mathcal{M}_{p},\mathcal{K}, \mathcal{D}, g_l$ & Eq~\eqref{reward:ESPL}\\
    \end{tabular}
    \caption{Inputs, policies and rewards for learning-based agents. Additionally, all agents'  input consists of $\mathcal{M}, \mathcal{M}_o, \mathcal{M}_a, \phi$ (where $\phi$ is the orientation as defined in Section~\ref{subsec:dataset}).}
    \label{tab:statespace}
\end{table}

\subsubsection{Hyperparameters}
\label{HyperParameters}
We train agents with Rllib implementation of DD-PPO algorithm. Each agent is trained for 100M steps with 8x Tesla V100 16GB GPUs and on a 64-core Intel Xeon E5-2686 CPU. Table~\ref{tab:hyperparam} shows the training and evaluation hyperparameters.
\begin{table}[h!]
    \rowcolors{1}{gray!3}{gray!7}
    \centering
    \begin{tabular}{c|c}
        Symbol & Value\\ \hline
        $\theta$, $r_s$, ${max}_t$, $n_o$, $n_r$, $c$ & $360^{\circ}$, 2.0m, 100, 5, 3, 3\\
        batch size & 128\\
        discount factor $\gamma$ & 0.9\\
        lambda $\lambda$ & 1.0\\
        vf-loss-coeff & 0.8\\
        entropy-coeff & $10^{-1}$ to $10^{-3}$ in 5M steps\\
        vf-clip-param & 100\\
        clip-param $\varepsilon$& 0.1\\
    \end{tabular}
    \caption{Training and evaluation hyperparameters for benchmarking unless otherwise specified.}
    \label{tab:hyperparam}
\end{table}

\subsubsection{Comparison of learnt exploration methods with frontier-based exploration methods}

Table~\ref{tab:exploreonlyresults} shows a detailed comparison of learnt exploration agents with WFBE agents for purely exploration task on MORP test scenes. LE agents were also trained for the exploration task. The goal of the exploration task is to find all objects in a given scene.

\begin{table}[h]
    \fontsize{9pt}{10pt} \selectfont
    \centering
    \rowcolors{1}{gray!3}{gray!7}
    \begin{tabular}{c | cccc||c|c}
        \multicolumn{1}{c|}{} &
        \multicolumn{5}{c}{Path Length $l(m)$}\\
        \hline
        \multicolumn{1}{c|}{\diagbox{Metds.}{$n_o$}} & \multicolumn{1}{c|}{$1$} & \multicolumn{1}{c|}{$3$} & \multicolumn{1}{c|}{$5$} & \multicolumn{1}{c||}{$10$} & \multicolumn{1}{c|}{$avg$} & \multicolumn{1}{c}{$FO$}\\
        \hline
        {RND} & 368 & 870 & 1250 & 2053 & 1104 & 81\\
        {WFBE$_\text{1}$} & 198 & 336 & 393 & 474 & 346 & 84\\
        {WFBE$_\text{0.5}$} & 146 & 268 & 339 & 440 & 293 & 58\\
        {WFBE$_\text{0}$} & 183 & 349 & 440 & 601 & 386 & 71\\
        {WFBE$_\text{r}$} & 146 & 260 & 320 & 411 & 280 & 59\\
        {LE$_\text{w}$} & 146 & 267 & 338 & 440 & 293 & 92\\
        {LE$_\text{disc}$} & 176 & 341 & 425 & 548 & 366 & 70\\
    \end{tabular}
    \caption{Comparison of exploration methods: We evaluate exploration agents using exploration only task  on MORP test scenes. We compare agent's path length required to find all objects in the scene and path length until the first object is found ($FO$).}
    \label{tab:exploreonlyresults}
\end{table}


\end{document}